# Differentiating Features for Scene Segmentation Based on Dedicated Attention Mechanisms

Zhiqiang Xiong[1], Zhicheng Wang[1], Zhaohui Yu[1], Xi Gu[1]

[1] Research Center of CAD, Tongji University, Shanghai, People's Republic of China
{zq_xiong, Zhichengwang, 1833017, 1833015}@tongji.edu.cn

## Abstract

*Semantic segmentation is a challenge in scene parsing. It requires both context information and rich spatial information. In this paper, we differentiate features for scene segmentation based on dedicated attention mechanisms (DF-DAM), and two attention modules are proposed to optimize the high-level and low-level features in the encoder, respectively. Specifically, we use the high-level and low-level features of ResNet as the source of context information and spatial information, respectively, and optimize them with attention fusion module and 2D position attention module, respectively. For attention fusion module, we adopt dual channel weight to selectively adjust the channel map for the highest two stage features of ResNet, and fuse them to get context information. For 2D position attention module, we use the context information obtained by attention fusion module to assist the selection of the lowest-stage features of ResNet as supplementary spatial information. Finally, the two sets of information obtained by the two modules are simply fused to obtain the prediction. We evaluate our approach on Cityscapes and PASCAL VOC 2012 datasets. In particular, there aren't complicated and redundant processing modules in our architecture, which greatly reduces the complexity, and we achieving 82.3% Mean IoU on PASCAL VOC 2012 test dataset without pre-training on MS-COCO dataset.*

## 1. Introduction

Semantic segmentation is a task of classifying image at the pixel level, and segment scene into different areas with semantic classes. A few examples are shown in Figure 1. It can be widely applied to the fields of automatic driving, scene understanding.

The accuracy of semantic segmentation is affected not only by semantic classifications, but also by the location of classification label for pixels, which is reflected in the consistency within categories and the edge details of some objects. Recently, many methods based on Fully Convolutional Networks (FCNs) [1] are proposed to address above problems. On the one hand, to obtain abundant semantic information, context information at different stages of encoder is usually fused during decoder [2][3]. Some works [4][5][6] aggregate multi-scale context information generated by different dilated convolutions or different scale pooling operations, and some works [8] enlarge the kernel size to grab context information. But the context information is not equally important to segmentation, it should be selectively enhanced with the guidance of global information.

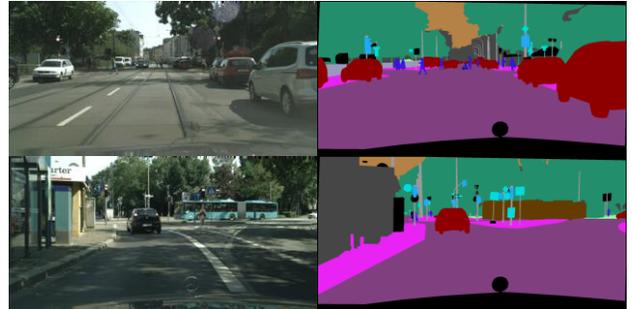

(a) Image  (b) Ground Truth

Figure 1. Illustration of complex scenes in Cityscapes dataset. The goal of image segmentation is to classify scenes at the pixel level, and segment image into different areas with semantic classes.

On the other hand, a large amount of spatial information is lost because of consecutive convolutions and pooling operations. [2][3] make up for the spatial information by integrating high-level and mid-level features. Some methods capture spatial information by increasing the receptive field for high-level features [4][5][6], but there is not much spatial information preserved in high-level features.

To address above problems, we propose a method called differentiating features for scene segmentation based on dedicated attention mechanisms (DF-DAM), that uses attention mechanisms in 2D-positions of low-level features and 1D-channels of high-level features, respectively. Both attention mechanisms are under the guidance of the information extracted by the encoder. For 1D-channel attention, we designed a Dual Attention Fusion Module (DAFM) to readjust the proportion of each feature map in the high-level features. Each channel can be regarded as a kind of feature, and different features have different effects on the results. Some channels represent common features of different categories, which have little or even negative effects on classification, while others are unique to different categories. For 2D-position attention module, we use the features fused by DAFM to filter the low-level features with spatial information. As a result of consecutive convolutions and pooling operations, high-level features lack spatial information and perform poorly in edge of objects. On the contrary, low-level features retain a lot of spatial information, which can make up for the defects of high-level features. However, the context information of low-level features is insufficient to guarantee the internal consistency of segmentation objects. Therefore, useful spatial information in low-level features is selected and unhelpful features is discarded.

Our main contributions can be summarized as follows:
- We explicitly differentiate the features of encoder, and propose a framework to process the encoded high-level features and low-level features respectively to generate segmentation prediction.
- A 2D-position attention module is proposed to select the low-level features with affluent spatial information, and a Dual Attention Fusion Module with dual channel attention to weight different feature maps of high-levels.
- We prove the validity of our method on Cityscapes [24] and PASCAL VOC 2012 [25] datasets.

## 2. Related work

Recently, lots of methods based on FCNs [1] have made significant progress on different benchmarks of the semantic segmentation task. Most of them are designed to fuse the features of adjacent encoder layers for sufficient information.

**Spatial information:** FCN-based models obtain high-level semantic information by convolutional neural network (CNN) [10] with convolution and down-sampling pooling. However, high-level semantic information is not enough for pixel-level semantic segmentation tasks, the spatial information is essential for the details of segmentation. For more spatial information, Global Convolutional Network (GCN) [8] adopts "large kernel" to increase receptive field. PSPNet [4] uses multi-scales pooling to preserve the spatial information of the feature maps, while DUC [11], DeepLab-v2 [5], and DeepLab-v3 [9] uses multi-scales dilated convolution.

**Context information:** Context information is crucial for distinguishing seemingly similar classes. [4][9][12] use global average pooling to supplement global context information. [4][9][11][15] capture and merge different levels of context information by adding different receptive field.

**Encoder-decoder:** Encoder of the FCNs-based models extracted different levels of features, but too much spatial information is corrupted by the convolution and pooling operations. Some methods based on U-shape structure integrate these features to recover spatial information and refine the prediction with different decoders. For example, U-net [2] uses the skip connection, while RifineNet [3] utilizes Multi-Path Refinement structure to optimize prediction results. SegNet [13] adds pooling indices in the decoder to retain the details, and LRR [14] employs the Laplacian Pyramid Reconstruction network. However, the lost spatial information cannot be recovered easily.

**Attention mechanism:** Powerful deep neural network can encode lots of information, and attention mechanism can act as a leap-forward guide to screen the information [16][17][18][19][20][21]. In SENet [16], features were used to learn attention to revise themselves. DFN [17] learn the global context to filter features. [18] uses the attention mechanism on the size of the input images.

## 3. Our Method

In this section, we first introduce our method detailedly. Then, we elaborate the design details of the two attention modules. Finally, we describe a complete network architecture for scene segmentation.

### 3.1. Overview

For an image to be segmented, encoder (such as ResNet [22] and VGG [23]) composed of a series of convolution and pooling operations is usually used to capture the information in the image. However, information in different stages plays different roles in segmentation. To take full advantage of the proprietary nature of these information, we processed the information of different stages with different strategies.

As illustrated in Figure 2, we employ a pre-trained residual network as the backbone. We firstly divide the feature maps of ResNet into two groups: the feature maps of the lowest stage as spatial information and the feature maps of the highest two stages as context information. We propose a Dual Attention Fusion Module (DAFM) to fuse the context information. During the fusion of context information, the feature maps of the higher stage are enlarged to keep consistent with the size of the feature maps of the lower stage. Then, the fused information and spatial information will be fed into the 2D Position Attention Module (2DPAM). In 2DPAM, context information helps generate a positional attention matrix. 2DPAM and DAFM will output weighted spatial information and weighted context information, respectively. Finally, we aggregate the two sets of information to obtain the final prediction.

### 3.2. Dual Attention Fusion Module

During the fusion of feature maps of adjacent stages, the difference between information of different stages should be considered, as well as the powerful semantic information of higher stage. Our Dual Attention Fusion Module (DAFM) uses high-level feature maps to change the weight of low-level feature maps and its own weight. Specifically, we use the high-level features to learn two weight vectors, which are used to adjust the high-level features and the low-level features at the channel level, respectively. In detail, we enter the high-level features into two sets of identical structures, each using global average pooling to capture global information, and the weights are constrained between 0 and 1 by the sigmoid function. At the same time, we use $1 \times 1$ convolution to convert high-level features and low-level features to the features with same number of channels, which is the same as the number of weight vector channels. The two weight vectors are multiplied separately with the high-level features and low-level features, and then the weighted high-level and low-level features are added.

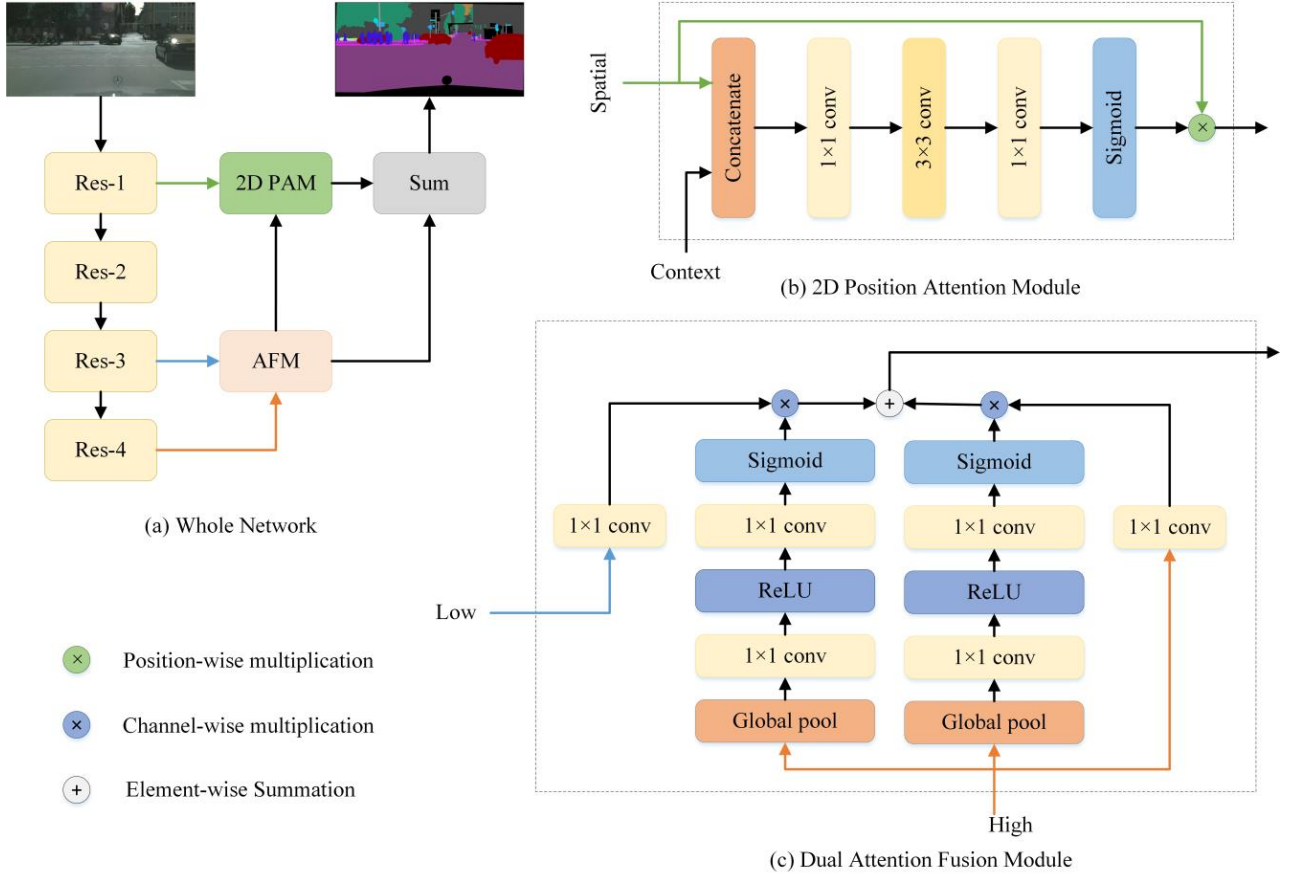

Figure 2. An overview of Our network. (a) Network Architecture. (b) Components of the 2D Position Attention Module (2DPAM). (c) Components of the Dual Attention Fusion Module (DAFM).

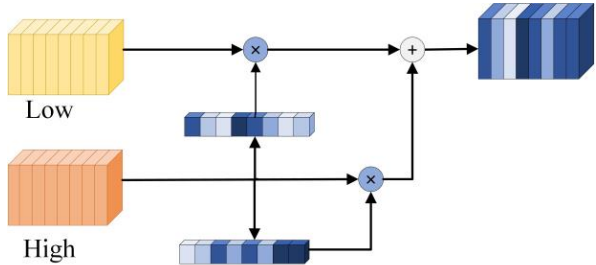

Figure 3. Dual Attention Fusion Module

As show in Figure 3, given two feature maps, Low and High, where $\{Low, High\} \in R^{C \times H \times W}$. For the $Low = X^L$, we assume that there are two positions, $A(i_A, j_A)$ and $B(i_B, j_B)$, whose values on channel $C_1$ differ by $\varepsilon_1^L$, and the values on channel $C_2$ differ by $\varepsilon_2^L$, where $\{(i_A, j_A), (i_B, j_B)\} \in D$, $D$ is the set of pixel positions.

$$\varepsilon_1^L = \left|X_{i_A, j_A, 1}^L - X_{i_B, j_B, 1}^L\right| \quad (1)$$
$$\varepsilon_2^L = \left|X_{i_A, j_A, 2}^L - X_{i_B, j_B, 2}^L\right| \quad (2)$$

Abstractly, $C_1$ and $C_2$ represent two different features, we assume that A and B need to maintain discrimination on $C_1$ and remain consistency on $C_2$. This means that $\varepsilon_1^L$ should be kept and $\varepsilon_2^L$ weakened. To address the problem, tow parameter $\alpha^L \in R^{C \times 1 \times 1}$ and $\alpha^H \in R^{C \times 1 \times 1}$ was introduced to adjust high-level and low-level features, where $\alpha = Sigmiod(X; \omega)$.

$$\overline{\varepsilon_1^L} = \left|\alpha_1^L X_{i_A, j_A, 1}^L - \alpha_1^L X_{i_B, j_B, 1}^L\right| = \alpha_1^L \varepsilon_1^L \quad (3)$$
$$\overline{\varepsilon_2^L} = \left|\alpha_2^L X_{i_A, j_A, 2}^L - \alpha_2^L X_{i_B, j_B, 2}^L\right| = \alpha_2^L \varepsilon_2^L \quad (4)$$

where the $\overline{\varepsilon_1^L}$ and $\overline{\varepsilon_2^L}$ are the weighted differences. The goal is achieved by increasing $\alpha_1^L$ and decreasing $\alpha_2^L$. For the feature maps obtained by adding $X^L$ and $X^H$, $\varepsilon_1$ and $\varepsilon_2$ represent the differences of A and B on channel $C_1$ and $C_2$.

$$\varepsilon_1 = \varepsilon_1^L + \varepsilon_1^H \quad (5)$$
$$\varepsilon_2 = \varepsilon_2^L + \varepsilon_2^H \quad (6)$$
$$\overline{\varepsilon_1} = \alpha_1^L \varepsilon_1^L + \alpha_1^H \varepsilon_1^H \quad (7)$$
$$\overline{\varepsilon_2} = \alpha_2^L \varepsilon_2^L + \alpha_2^H \varepsilon_2^H \quad (8)$$

In fact, the features of adjacent stages behave similarly at some positions, we can assume that $\varepsilon_1^L$ is equal to $\varepsilon_1^H$. If only one of $\alpha_1^L$ and $\alpha_2^L$ is applied, the range of adjustment is roughly halved. Therefore, the use of $\alpha^L$ and $\alpha^H$ can help adjust the effect of different features at different pixel positions, or obtain more discriminative features.

3.3. 2D Position Attention Module

With a large number of convolution and pooling operations, the high-level features obtain enough context information to make the overall judgment of the object more accurate, but the resolution reduction loses a lot of spatial information, resulting in unclear boundaries of the segmentation. In contrast, low-level features preserved high resolution and rich spatial information. However, if low-level features are directly combined with high-level features, the lack of low-level feature context information will affect the advantages of high-level features. We design a 2D Position Attention Module (2DPAM) to select useful spatial information in low-level features.

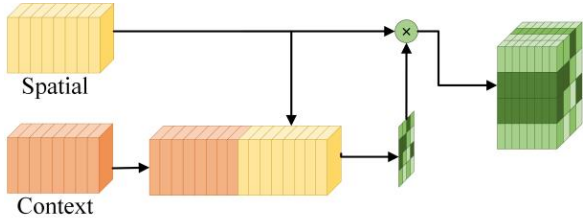

Figure 4. 2D Position Attention Module

As show in Figure 4, we concatenate low-level features (spatial information) and the features fused by DAFM at first, then the concatenated features is fed into the network of convolution and activation functions to obtain a confidence map. The low-level features and the confidence map are multiplied to get the filtered spatial information.

$$\overline{X^{SI}} = \beta X^{SI} = \begin{bmatrix} \beta_{1,1} X_{1,1}^{SI} & \cdots & \beta_{1,W} X_{1,W}^{SI} \\ \vdots & \ddots & \vdots \\ \beta_{H,1} X_{H,1}^{SI} & \cdots & \beta_{H,W} X_{H,W}^{SI} \end{bmatrix} \quad (9)$$

where $X^{SI} \in R^{C \times H \times W}$ is the spatial information (low-level features), $\beta \in R^{1 \times H \times W}$ is the confidence map.

We assume that $X_{i_P,j_P}^{SI}$ and $X_{i_P,j_P}^{CI}$ are the spatial and context features at the position $P(i_P, j_P)$, respectively. We introduce $\beta_{i_P,j_P}$ to affect the fusion of $X_{i_P,j_P}^{SI}$ and $X_{i_P,j_P}^{CI}$.

$$X^F = \beta_{i_P,j_P} X_{i_P,j_P}^{SI} + X_{i_P,j_P}^{CI} = \begin{bmatrix} \beta_{i_P,j_P} X_{i_P,j_P,1}^{SI} + X_{i_P,j_P,1}^{CI} \\ \vdots \\ \beta_{i_P,j_P} X_{i_P,j_P,C}^{SI} + X_{i_P,j_P,C}^{CI} \end{bmatrix} \quad (10)$$

where $\beta_{i_P,j_P}$ is the confidence at position P. If the effect of $X_{i_P,j_P}^{SI}$ on $X_{i_P,j_P}^{CI}$ is positive, then $\beta_{i_P,j_P}$ is increased. Otherwise, in order to avoid the wrong distribution of $X_{i_P,j_P}^{SI}$ affecting the correct distribution of $X_{i_P,j_P}^{CI}$, it is necessary to reduce $\beta_{i_P,j_P}$. By generating confidence of the low-level features of each pixel position, low-level features can be selectively used to take full advantage of the features while reducing the effects of error information.

3.4. Network Architecture

In order to apply DAFM and 2DPAM to encoder, we adopt 1x1 convolution to convert the feature maps of stages to 128 channels. In the final Sum Fusion, we first add the output of DAFM and 2DPAM, and then restore the feature maps to the size of input with a simple refine block to get the prediction. Moreover, like the deep supervision [26], we add two auxiliary losses to supervise the output from spatial information and context information. All the loss functions are Softmax loss.

$$L(y; w) = SoftmaxLoss(y; w) \quad (11)$$
$$loss = L_p(y_p; w) + \lambda_c L_c(y_c; w) + \lambda_s L_s(y_s; w) \quad (12)$$

where the $y_s$ and $y_c$ are the output from spatial information and context information, respectively, the $L_s$ and $L_c$ are the auxiliary loss. And the $y_p$ is the prediction of network, the $L_p$ is the principal loss function, the $loss$ is the joint loss function. Furthermore, we use the parameters $\lambda_s$, $\lambda_c$ to balance the principal loss and auxiliary loss. The $\lambda_s$, $\lambda_c$ in this paper is equal to 0.1 and 0.4, respectively.

4. Experiments

We evaluate the proposed method on two public datasets: Cityscapes and PASCAL VOC 2012. We first introduce the datasets and the implementation details, then we investigate the effects of each module of our method. Finally, we report the results of our method on Cityscapes and PASCAL VOC 2012 datasets.

**Cityscapes** The Cityscapes is a large dataset of 5,000 fine annotated images for urban scenes segmentation. The dataset contains 30 classes, 19 classes of which are used for training and evaluation. There are 2,979 images for training, 500 images for validation and 1,525 images for testing. Each image has 2,048 × 1,024 pixels.

**PASCAL VOC 2012** The Pascal VOC 2012 is one of the most commonly used semantic segmentation datasets, which contains 20 object classes and one background. In the dataset, 1,464 images for training, 1,449 images for validation and 1,456 images for testing. We augment the original dataset with the Semantic Boundaries Dataset [27], resulting in 10,582 images for training.

### 4.1. Implementation Details

In experiments, we apply the ResNet series pre-trained on ImageNet dataset [29] as the backbone, including ResNet-18, ResNet-50, ResNet-101, and we implement our method based on Pytorch.

**Data augmentation:** We adopt random horizontal flip, mean subtraction and random scale on the input images in training process, the scales contains {0.75, 1.0 1.25, 1.5, 1.75, 2.0}. Besides, we randomly crop the input image, the crop size is $1024 \times 1024$ and $512 \times 512$ for Cityscapes and PASCAL VOC 2012, respectively.

**Training Details:** We use mini-batch stochastic gradient descent (SGD) [28]. Inspired by [5][9], we employ a "poly" learning rate policy where the current learning rate equals to the initial learning rate multiplying $\left(1 - \frac{iter}{max\_iter}\right)^{power}$ with power 0.9. For Cityscapes, we use SGD with batch size 8, initial learning rate 0.01, momentum 0.9 and weight decay $5e^{-4}$. For PASCAL VOC 2012, we use SGD with batch size 16, initial learning rate 0.001, momentum 0.9 and weight decay $1e^{-5}$.

### 4.2. Ablation Study

In this subsection, we will decompose the network to verify the effect of each module. We evaluate our method on the validation set of Cityscapes [24] and PASCAL VOC 2012 [25]. Based on ResNet, we add DAFM and 2DPAM to fuse the first, third and fourth stages features. As a comparison, we use summation replace our module to build the baseline. To verify the effects of our modules, we conduct experiments with different setting in Table 1 on PASCAL VOC 2012 validation set, and Table 2 on Cityscapes validation set.

**PASCAL VOC 2012:** As show in Table 1, the modules significantly improved the performance. Compare to the baseline (ResNet-50), the use of Dual Attention Fusion Module raise the Mean IoU from 68.2% to 73.7%, an improvement of 5.5%. On this basis, the addition of 2D Position Attention Module further improves the Mean IoU to 74.6%. With a deeper backbone, the two modules improve the Mean IoU over baseline (ResNet-101) by 5.0%.

**Cityscapes:** As show in Table 2, the Dual Attention Fusion Module brings a 2.5% improvement over baseline (ResNet-18), and the employing of 2D Position Attention Module further improves the performance to 73.1%.

As show in Figure 5 and Figure 6., with the DAFM, some misclassification within the objects was eliminated, such as the 'car' in the first row, the 'sidewalk' in the second row of Figure 6., and the 'dog' in the third row of Figure 5. Furthermore, the 2DPAM filters out the wrong information in the low-level features while preserving the spatial information that has a positive effect, which makes the segmentation more accurate and more holistic.

| Backbone | Method | Mean IoU% |
|---|---|---|
| Res50 | Sum (baseline) | 68.2 |
| Res50 | DAFM | 73.7 |
| Res50 | DAFM + 2DPAM | 74.6 |
| Res101 | Sum (baseline) | 72.9 |
| Res101 | DAFM | 76.8 |
| Res101 | DAFM + 2DPAM | 77.9 |

Table 1. Ablation study for modules on PASCAL VOC 2012 val set. *DAFM* represents Dual Attention Fusion Module, *2DPAM* represents 2D Position Attention Module.

| Backbone | Method | Mean IoU% |
|---|---|---|
| Res18 | Sum (baseline) | 70.3 |
| Res18 | DAFM | 72.8 |
| Res18 | DAFM + 2DPAM | 73.1 |

Table 2. Ablation study for modules on Cityscapes val set.

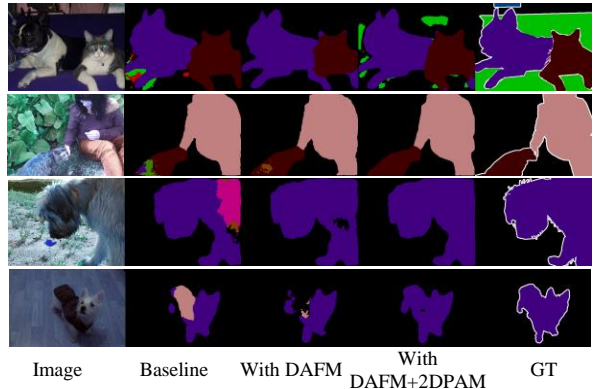

Image    Baseline    With DAFM    With DAFM+2DPAM    GT

Figure 5. Examples of our results on PASCAL VOC 2012 dataset.

### 4.3. Visualization and Analysis of Attention

To illustrate the effect of the attention mechanism in the network explicitly, we visualized the results of the two attention modules.

For the Dual Attention Fusion Module, the features of high-level and low-level to be fused are 128 channels, and feature maps of each channel are in size of $H \times W$. Therefore, the two vectors used to adjust the features of the high-level and low-level are in dimensions of 128. To analyze the meaning of the channel weights, we visualize several channel maps of the low-level features to an image with a size of $H \times W$. As show in Figure 8, $51^{th}$ channel map in the second column with a low weight and $11^{th}$

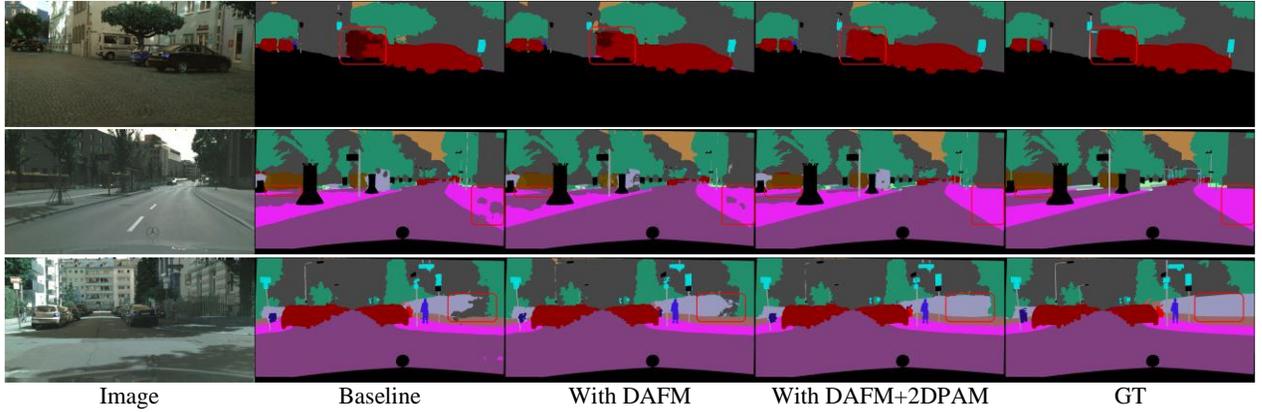
Figure 6. Examples of our results on Cityscapes dataset.

channel map in the third, $126^{th}$ channel map in the fourth column with a high weight. The features in the $51^{th}$ channel map is difficult to distinguish, since most pixel position have similar values. As a comparison, there are discriminative features in the $11^{th}$ and $126^{th}$ channel map, some of the segmentation objects can be clearly distinguished, such as the car, the boundary of the tree in the $11^{th}$ channel map, and the trees, road in the $126^{th}$ channel map. That is to say, the $11^{th}$ and $126^{th}$ channel represent some kind of abstract features, which can distinguish the objects with these features from other objects. The above explained why the attention mechanism gives a high weight to $11^{th}$ and $126^{th}$ channel, and a low weight to the $51^{th}$ channel.

Further, we visualize the weight vectors of the high-level and low-level features of the two examples in the Figure 8. The Figure 7(a) is the weight of low-level and high-level features of the image in the first row of Figure 8, while the Figure 7(b) is the weight of low-level features of the first two images in Figure 8. From the Figure 7 we can learn that: 1) the weights of channels are different in different images, which indicates that the convolution operation cannot meet the requirements of feature weight adjustment, so it is useful to introduce the channel attention mechanism; 2) as show in the Figure 7 (b), the weights distribution of the low-level features of different images are roughly the same, which implies that channels specifically represent some abstract features, and each feature is of different importance; 3) as show in the Figure 7 (a), the weight of high-level features is generally higher than the weight of the low-level features, and the distribution of them is very different, the weight of high-level is dominant.

For the 2D Position Attention Module, we visualize the confidence map of spatial information. As show in the five column of Figure 8, the brighter the pixel position represents the more useful and credible spatial information. It can be clearly seen that the highlights are almost at the boundaries of the segmentation objects. It shows that, the spatial information lacking in the high-level features is

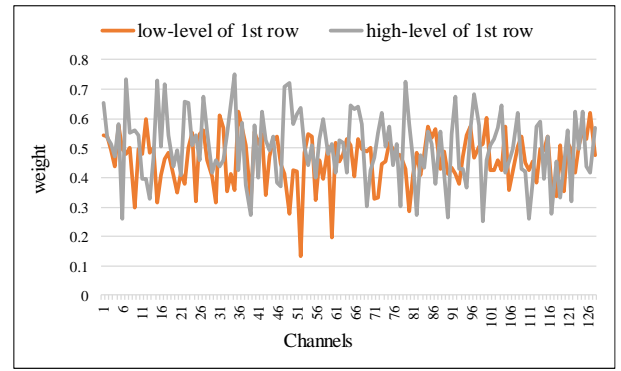
(a)

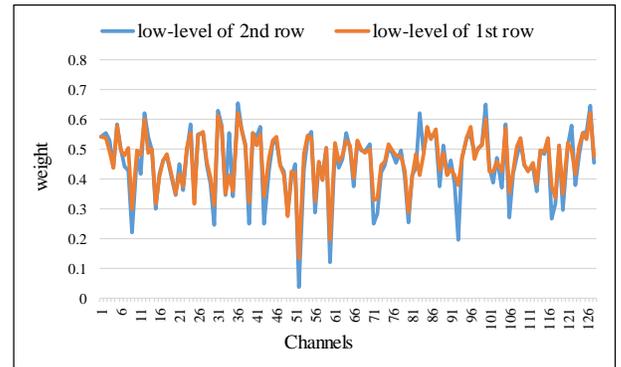
(b)

Figure 7. Channel weight vectors of the first two examples in the Figure 8. (a) the weight of high-level features and low-level features of the first example. (b) the weight of the low-level features of the first two example.

preserved by confidence map, while the context information in low-level features is not enough, so the part lacking the semantic information is filtered.

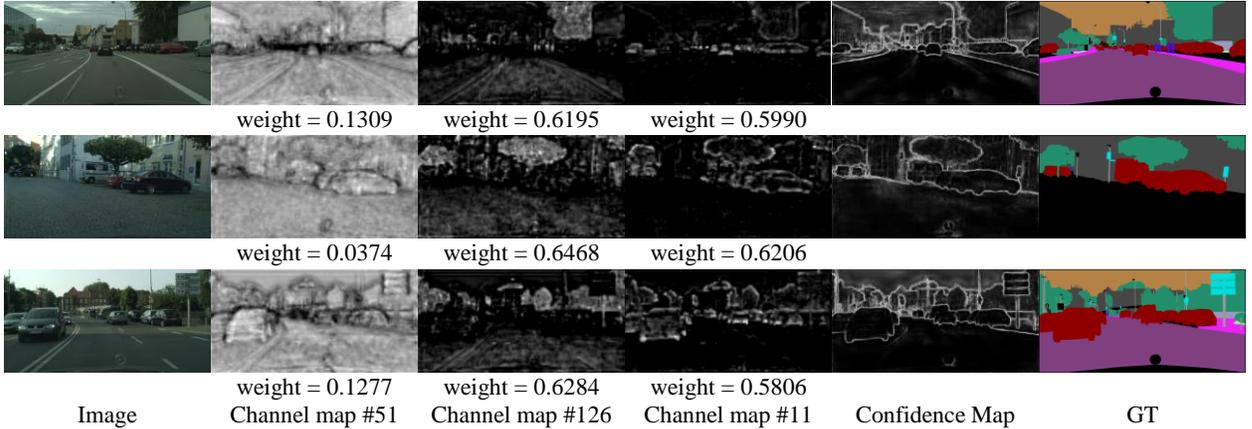

| | | | | | |
|---|---|---|---|---|---|
| | weight = 0.1309 | weight = 0.6195 | weight = 0.5990 | | |
| | weight = 0.0374 | weight = 0.6468 | weight = 0.6206 | | |
| | weight = 0.1277 | weight = 0.6284 | weight = 0.5806 | | |
| Image | Channel map #51 | Channel map #126 | Channel map #11 | Confidence Map | GT |

Figure 8. Visualization results of attention modules on Cityscapes val set. For each row, we show an input image, and three channel maps from the low-level features, where the maps form $51^{th}$, $126^{th}$, $11^{th}$, respectively, as well as the confidence map of the 2D Position Attention Module. Finally, corresponding ground truth are provided.

| Method | MS | Flip | Mean IoU% |
|---|---|---|---|
| Ours (Res18) | | | 73.1 |
| Ours (Res18) | √ | | 74.6 |
| Ours (Res18) | √ | √ | 74.9 |

Table 3. Ablation study for improvement strategies on Cityscapes val set. *MS* represent Multi-Scale, and *Flip* represent left-right flip.

| Method | Params | GFLOPs | Mean IoU% |
|---|---|---|---|
| FCN-8s-heavy [1] | 134.46 | 188.93 | 65.3 |
| Dilation10 [15] | 134.35 | 134.33 | 67.1 |
| DeepLab(Res101) [5] | 43.90 | 148.73 | 70.4 |
| RefineNet(Res101) [3] | 118.01 | 659.14 | 73.6 |
| DUC(Res101) [11] | 148.04 | 95.87 | 77.6 |
| PSPNet(Res101) [4] | 75.04 | 230.01 | **78.4** |
| Ours(ResNet18) | **11.57** | **10.31** | 73.6 |

Table 4. Comparison of our approach with some representative method on Cityscapes test set. *GFLOPs* are estimated for input of $3 \times 640 \times 320$.

| Method | MS | Flip | Mean IoU% |
|---|---|---|---|
| Ours (Res101) | | | 77.90 |
| Ours (Res101) | √ | | 79.55 |
| Ours (Res101) | √ | √ | 79.83 |

Table 5. Ablation study for improvement strategies on PASCAL VOC 2012 val set.

| Method | Params | GFLOPs | Mean IoU% |
|---|---|---|---|
| FCN-8s-heavy [1] | 134.49 | 195.03 | 67.2 |
| DeepLab(Res101) [5] | 44.05 | 149.22 | 71.6 |
| RefineNet(Res101)[+] [3] | 118.01 | 659.24 | 82.4 |
| DUC(Res152)[+] [11] | 163.68 | 111.30 | **83.1** |
| PSPNet(Res101) [4] | 75.05 | 230.02 | 82.6 |
| Ours(ResNet101) | **43.24** | **35.52** | 82.3 |

Table 6. Comparison of our approach with some representative method on PASCAL VOC 2012 test set. Methods pre-trained on MS-COCO are marked with '+'.

### 4.4. Results on Cityscapes

In evaluation, following [17][30], we adopt the multi-scale input with scales = {0.5, 0.75, 1.0, 1.25, 1.5, 1.75, 2.0} and left-right flip on the image, and we training our method with only fine data of Cityscapes. As show in Table 3, the multi-scale input improved the Mean IoU by 1.5% to 74.6%, and the left-right flip bring an improvement by 0.3%. We trained our network with the best setting, and experiment on Cityscapes test dataset [24]. Furthermore, we compare our approach with other representative methods, including Params, GFLOPs, and Mean IoU. As show in Table 4, our method yields Mean IoU 73.6% on Cityscapes test set. Compare with other methods, our method greatly reduces the parameters and the GFLOPs. Our method is 4.8% less accurate than PSPNet [4] while the learning parameters is $6.5\times$ fewer, GFLOPs is $22\times$ fewer. Compared to RefineNet [3] with the same Mean IoU, our learning parameters is $10\times$ fewer, and GFLOPs is $64\times$ fewer. RefineNet uses multiple identical refine blocks to fuse multi-scale features, which complicates the network, while our approach adopts two dedicated attention mechanisms to adjust features of different phases for complementary fusion, making the network lightweight.

| Methods | mIoU | aero | bike | bird | boat | bottle | bus | car | cat | chair | cow | table | dog | horse | mbike | person | plant | sheep | sofa | train | tv |
|---|---|---|---|---|---|---|---|---|---|---|---|---|---|---|---|---|---|---|---|---|---|
| FCN-8s [1] | 67.2 | 82.4 | 36.1 | 75.6 | 61.5 | 65.4 | 83.4 | 77.2 | 80.1 | 27.9 | 66.8 | 51.5 | 73.6 | 71.9 | 78.9 | 77.1 | 55.3 | 73.4 | 44.3 | 74.0 | 63.2 |
| Zoom-out [31] | 69.6 | 85.6 | 37.3 | 83.2 | 62.5 | 66.0 | 85.1 | 80.7 | 84.9 | 27.2 | 73.2 | 57.5 | 78.1 | 79.2 | 81.1 | 77.1 | 53.6 | 74.0 | 49.2 | 71.7 | 63.3 |
| ParseNet [12] | 69.8 | 84.1 | 37.0 | 77.0 | 62.8 | 64.0 | 85.8 | 79.7 | 83.7 | 27.7 | 74.8 | 57.6 | 77.1 | 78.3 | 81.0 | 78.2 | 52.6 | 80.4 | 49.9 | 75.7 | 65.0 |
| DeepLab [5] | 71.6 | 84.4 | 54.5 | 81.5 | 63.6 | 65.9 | 85.1 | 79.1 | 83.4 | 30.7 | 74.1 | 59.8 | 79.0 | 76.1 | 83.2 | 80.8 | 59.7 | 82.2 | 50.4 | 73.1 | 63.7 |
| Piecewise [32] | 75.3 | 90.6 | 37.6 | 80.0 | 67.8 | 74.4 | 92.0 | 85.2 | 86.2 | 39.1 | 81.2 | 58.9 | 83.8 | 83.9 | 84.3 | 84.8 | 62.1 | 83.2 | 58.2 | 80.8 | 72.3 |
| LRR-CRF [14] | 75.9 | 91.8 | 41.0 | 83.0 | 62.3 | 74.3 | 93.0 | 86.8 | 88.7 | 36.6 | 81.8 | 63.4 | 84.7 | 85.9 | 85.1 | 83.1 | 62.0 | 84.6 | 55.6 | 84.9 | 70.0 |
| DANet [33] | 82.6 | - | - | - | - | - | - | - | - | - | - | - | - | - | - | - | - | - | - | - | - |
| PSPNet [4] | 82.6 | 91.8 | **71.9** | 94.7 | 71.2 | 75.8 | 95.2 | 89.9 | **95.9** | **39.3** | 90.7 | 71.7 | 90.5 | **94.5** | 88.8 | **89.6** | **72.8** | 89.6 | **64.0** | 85.1 | **76.3** |
| EncNet [34] | **82.9** | 94.1 | 69.2 | **96.3** | **76.7** | **86.2** | **96.3** | 90.7 | 94.2 | 38.8 | **90.7** | **73.3** | 90.0 | 92.5 | 88.6 | 87.9 | 68.7 | 92.6 | 59.0 | 86.4 | 73.4 |
| Ours | 82.3 | **95.6** | 70.7 | 91.1 | 73.0 | 76.1 | 94.1 | **91.3** | 93.7 | 37.8 | 90.2 | 72.1 | **91.0** | 93.0 | **90.6** | 88.7 | 72.3 | **92.6** | 59.7 | **86.6** | 72.5 |

Table 7. Per-class results on PASCAL VOC 2012 test set without pre-training on MS-COCO dataset.

### 4.5. Results on PASCAL VOC 2012

As in the experiment on Cityscapes test set, we employ multi-scale input and left-right flip in evaluation on PASCAL VOC 2012 test set [25], as show in Table 5. Furthermore, since the PASCAL VOC 2012 dataset provides higher quality of annotation than the augmented datasets [27], we fine-tune our model on PASCAL VOC 2012 trainval set for evaluation on the test set. As show in Table **6**, our method achieves 82.3% [1] Mean IoU on PASCAL VOC 2012 test set without pre-training on MS-COCO dataset [35], detailed results are listed in Table 7. Our method is 0.7 % less Mean IoU than DUC who trained their model with extra MS-COCO dataset and employed a deeper backbone (ResNet152). With the same backbone, our method is 0.2% less Mean IoU than PSPNet [4], but our GFLOPs are much less than PSPNet (only about 1/7 of PSPNet).

### 5. Conclusion

In this paper, we proposed an effective network called differentiating features for scene segmentation based on dedicated attention mechanisms (DF-DAM). Specifically, two attention modules are introduced to optimize the high-level and low-level features in the feature extraction network, respectively, and the optimized features are used as the source of context information and spatial information. We demonstrated that the two modules can improved the segmentation performance remarkably by ablation experiments, and visually analyzed the intermediate features, which verified the effects of the two attention mechanisms in the architecture. In particular, our method aims at differentiation of the features, which has no complicated and repetitive structure, such as refine block in RefineNet, while ensuring access to sufficient information, the complexity of the model is greatly reduced. Our method was evaluated on the Cityscapes and PASCAL VOC 2012 dataset, and achieved great accuracy.

---

[1] The result link to the VOC evaluation server:
http://host.robots.ox.ac.uk:8080/anonymous/O5UYFF.html